\documentclass[11pt]{article}
\usepackage{coling2020}
\usepackage{amsmath}
\usepackage [autostyle, english = american]{csquotes}
\usepackage{tabularx}
\usepackage{url}

\usepackage{times}
\usepackage{latexsym}

\usepackage{algorithm}
\usepackage{algorithmic}
\usepackage{graphicx}
\usepackage{hyperref}
\usepackage{diagbox}

\MakeOuterQuote{"}

\newcommand{\figref}[1]{Figure~\ref{fig:#1}}
\newcommand{\tabref}[1]{Table~\ref{tab:#1}}

\newcommand{\secref}[1]{Section~\ref{sec:#1}}

\renewcommand{\eqref}[1]{Equation~\ref{eq:#1}}

\newcommand{\algref}[1]{Algorithm~\ref{alg:#1}}

\title{Mischief: A Simple Black-Box Attack Against Transformer Architectures}
\author{Adrian de Wynter \\
  Amazon Alexa \\
  \tt{dwynter@amazon.com} \\}

\begin{document}
\maketitle

\begin{abstract}
We introduce Mischief, a simple and lightweight method to produce a class of human-readable, realistic adversarial examples for language models.
We perform exhaustive experimentations of our algorithm on four transformer-based architectures, across a variety of downstream tasks, as well as under varying concentrations of said examples. 
Our findings show that the presence of Mischief-generated adversarial samples in the test set significantly degrades (by up to 20\%) the performance of these models with respect to their reported baselines. 
Nonetheless, we also demonstrate that, by including similar examples in the training set, it is possible to restore the baseline scores on the adversarial test set. 
Moreover, for certain tasks, the models trained with Mischief set show a modest increase on performance with respect to their original, non-adversarial baseline.
\end{abstract}

\section{Introduction}

An adversarial attack on deep learning systems, as introduced by \newcite{Szegedy} and \newcite{Goodfellow}, consists on any input that may be designed explicitly to cause poor performance on a model. They are traditionally split in two major categories: white-box and black-box. In the former, the adversarial inputs are found--or rather, learned--through a perturbation of the gradient. The latter, on the other hand, assumes that there is no access to the model's gradient, and thus said adversarial examples are often found by trial-and-error.

In computer vision, such attacks typically involve injecting learned noise into small areas of the input image. This noise is unnoticeable to a human user, but just about complex enough to cause the network to fail to perform as expected. 
In contrast, for text-based systems, this noticeability-versus-failure tradeoff is not as clear. Since machine learning-based language models embed an input string into a vector space for further processing in other tasks, from a feasibility point of view it would be more realistic to determine which changes in the string, and not the vector, are more likely to negatively affect the model.

In an attempt to make our work more readily applicable to existent systems, we concentrate ourselves solely on such black-box attacks. Moreover, we focus mostly on architectures that leverage the transformer layer from \newcite{Attention}, such as BERT \cite{BERT}, as their high performance in multiple language modeling tasks makes them ubiquitous in both research and production pipelines.

In this work we present Mischief, a procedure that generates a class of such adversarial examples. In order to remain within our constraints, Mischief leverages a well-known phenomenon from psycholinguistics first described by \newcite{RawlinsonThesis}. We characterize the impact of our algorithm on the performance of four selected transformer-based architectures, by carrying out exhaustive experiments across a variety of tasks and concentrations of adversarial examples.

Our experimentation shows that the presence of Mischief-generated examples is able to significantly downgrade the performance of the language models evaluated. 
However, we also demonstrate that, at least for the architectures evaluated, including Mischief-generated examples into the training process allows the models to regain, and sometimes increase, their baseline performance in a variety of downstream tasks.

\section{Related Work}\label{sec:relatedwork}

Adversarial attacks in the context of learning theory were perhaps first described by \newcite{KearnsLi}. However, such examples \emph{per se} predate machine learning (e.g., techniques to circumvent spam filters) by a wide margin \cite{ollmann2007phishing}. 
On the other hand, the study of adversarial attacks on deep neural networks, albeit relatively recent, fields a large number of important contributions in addition to the ones mentioned on the introduction, and it is hard to name them all. However, an excellent introduction on this topic, along with historical notes, can be found in \newcite{Battista}. 
In the context of Natural Language Understanding (NLU), the work by \newcite{JiaAndLiang} was arguably the first where such notions were formally applied to the intersection of language modeling and deep learning. Moreover, the well-known research by \newcite{Ebrahimi}, \newcite{BelinkovBisk}, and \newcite{Minervini} showed that the large majority of existing language models are extremely vulnerable to both black-box and white-box attacks. Indeed, the Mischief algorithm is similar to that one of \newcite{BelinkovBisk} with variations on noise levels, and applied over a wider range of natural language tasks.

Nonetheless, the large majority of the procedures presented in such papers were often considered to be unrealistic \cite{NLUSurvey}, and it wasn't until \newcite{Pruthi} and \newcite{HotFlip} where more practical attack and defense mechanisms were introduced. This paper is more closely aligned to theirs, although it differs in key aspects regarding contributions and methodology. 
Regardless, our work is meant to add to this body of research, with a specific focus on black box-based sentence level attacks for transformer architectures. 
The interested reader can find a more comprehensive compendium of the history of adversarial attacks for NLU in the survey by \newcite{NLUSurvey}.

We elaborate on a few studies of the research first reported by Graham Rawlinson \cite{Rawlinson} in \secref{main}. In addition to these papers, it is important to point out that there are quite a few papers around this phenomenom. For example, the work by \newcite{Perea2003} and \newcite{CASINO} expanded upon said research by exploring other types of permutations; while \newcite{Gomez} and \newcite{Norris} attempted to explain this phenomenon from a statistical perspective, and an analysis of some of the leading theories around positional encoding can be found in the articles by \newcite{Davis2006ContrastingFD} and \newcite{WhitneySeriol}. Finally, a compilation of the works around this effect can be found in \newcite{Davis}.

\section{Mischief}\label{sec:main}

\subsection{Background}\label{sec:background}

Graham Rawlinson described in his doctoral thesis \shortcite{RawlinsonThesis} a phenomenon where permuting the middle characters in a word, but leaving the first and last intact, had little to no effect on the ability of a human reader to understand it. 
It was shown in a few other studies that said permutation does tend to slow down readability \cite{Rayner2006}, and that the type of permutation (i.e., the position of the permuted substring) is relevant to the comprehension of the text \cite{Schoonbaert}, as well as any context \cite{Pelli2003} added. 

It could be argued that the act of shuffling the characters in an input word will have a naturally detrimental effect on any language model. Most models rely on a tokenizer and a vocabulary to parse the input string; thus, the presence of an adversarial example as an input to a pretrained model implies that the input will very likely be mapped to a low-information, or even incorrect, vocabulary element. 
On the other hand, the attention mechanism that lies at the heart of the transformer architecture does not have a concept of word order \cite{Attention}, and relies on statistical methods to learn syntax \cite{Peters}. 
It has also been shown to prefer, in some architectures, certain specific tokens and other coreferent objects \cite{clark-etal-2019-bert,kovaleva-etal-2019-revealing}. 
This suggests that, although models relying on these artifacts may be resilient to slight changes in the input, the right concentration of permuted words may lead to a degradation in performance--all while remaining understandable by a human reader.

Note that we do not alter the order of the words in the sentence, as that may risk losing a significant amount of semantic and lexical information, and thus it will no longer be considered a practical adversarial attack.

\subsection{The Mischief Algorithm}

We define the Generalized Rawlinson adversarial example (GRA) as a permutation $\sigma$ on a word $w = w_1,w_2,\dots,w_n$, where $n>3$, such that GRA$(w) = w_1,\sigma(w_2),\dots,\sigma(w_{n-1}),w_n$. The algorithm that we use to generate such examples, which we call Mischief, is a function that acts on a text corpus and takes in two parameters $p, r \in [0, 1]$. Here, $p$ denotes the proportion of the dataset to perform "Mischief" on, and $r$ is the probability of a word $w$ in a given line to be randomized; that is, to perform GRA$(w)$. An implementation of Mischief can be seen in \algref{mischief}.

\begin{algorithm}[tb]\caption{Simplified Mischief: The $+$ operator denotes string concatenation.}\label{alg:mischief}
\begin{algorithmic}
   \STATE {\bfseries Input:} dataset $D$, proportion $p$, probability $r$
   \FOR{sentence $s \in D$}

     \STATE Draw probability $\pi_s \sim P$
     \IF{$\pi_s \leq p$}
       \FOR{word $w \in s$}
         \STATE Draw probability $\pi_r \sim P$
         \IF{$\pi_r \leq r \land \vert w \vert > 3$}
           \STATE $w' = w_1 + \text{GRA}(w_{2:-2}) + w_{-1}$
           \STATE Replace $w \in s$ with $w'$
         \ENDIF
       \ENDFOR
     \ENDIF

   \ENDFOR
\end{algorithmic}
\end{algorithm}

\section{Experimentation}\label{sec:results}

We evaluate our approach with four transformer-based models: BERT (large, cased) \cite{BERT}, RoBERTa (large) \cite{RoBERTa}, XLM (2048-en) \cite{XLM}, and XLNet (large) \cite{XLNet}. All of them were selected due to their high performances on the Generalized Language Understanding Evaluation (GLUE)\footnote{\url{https://gluebenchmark.com/leaderboard}} benchmarks, which is a set of ten distinct NLU tasks designed to showcase the candidate's ability to model and generalize language \cite{GLUE}.

For every model and task, we apply four different concentrations $r=\{25\%, 50\%, 75\%, 100\%\}$ of Mischief on the training set; additionally, for each of these concentrations we test different combinations of Mischief-no Mischief on the test and training set, totaling $640$ distinct experiments. Due to the large complexity of the task, we maintain $p=1$ across all experiments.
A summary of our findings and general experimental set up is described in \tabref{summary}.

\setlength\extrarowheight{3pt}
\begin{table}[ht]
\centering
\small
\begin{tabularx}{\linewidth}{|c||X|X| } \hline
\backslashbox[20mm]{Training}{Test} & \textsc{Mischief} & \textsc{No Mischief} \\ \hline\hline
\textsc{No Mischief} & An adversarial attack: significant performance degradation. & Baseline from the GLUE benchmarks. \\ \hline
\textsc{Mischief}    & Proposed defense: minimal performance degradation. & Minimal degradation, or increased performance. \\ \hline
\end{tabularx}
\caption{Summary of our findings for all experiments, models, and variations.}\label{tab:summary}
\end{table}

\subsection{Methodology}

In order to obtain the performance of a model on a test set, an experimenter must first upload the raw predictions to the GLUE website. 
Due to the number of experiments we performed, along with our need to modify the test sets, we opted out from evaluating every result on the website. Instead, we treated the provided validation set as a test set, and generated a small validation set from a $90\%-10\%$ split of the training set.

It is well-known that the language models we tested are highly sensitive to initialization. Given that in most tasks we observed significant variation of results accross multiple experiments, we report the average result for ten random seeds, bringing up the total number of experiments to $6,\!400$. 
However, as done in \newcite{kovaleva-etal-2019-revealing}, we opt to not report the CoLA or WNLI benchmarks, as their small training set size made them remarkably sensitive to variations in the experimentation, and their inclusion could bias our summary results for the following sections.

\subsection{Effects of Mischief in the Test Set}\label{sec:testset}

\begin{figure}
\centering
\includegraphics[width=\columnwidth]{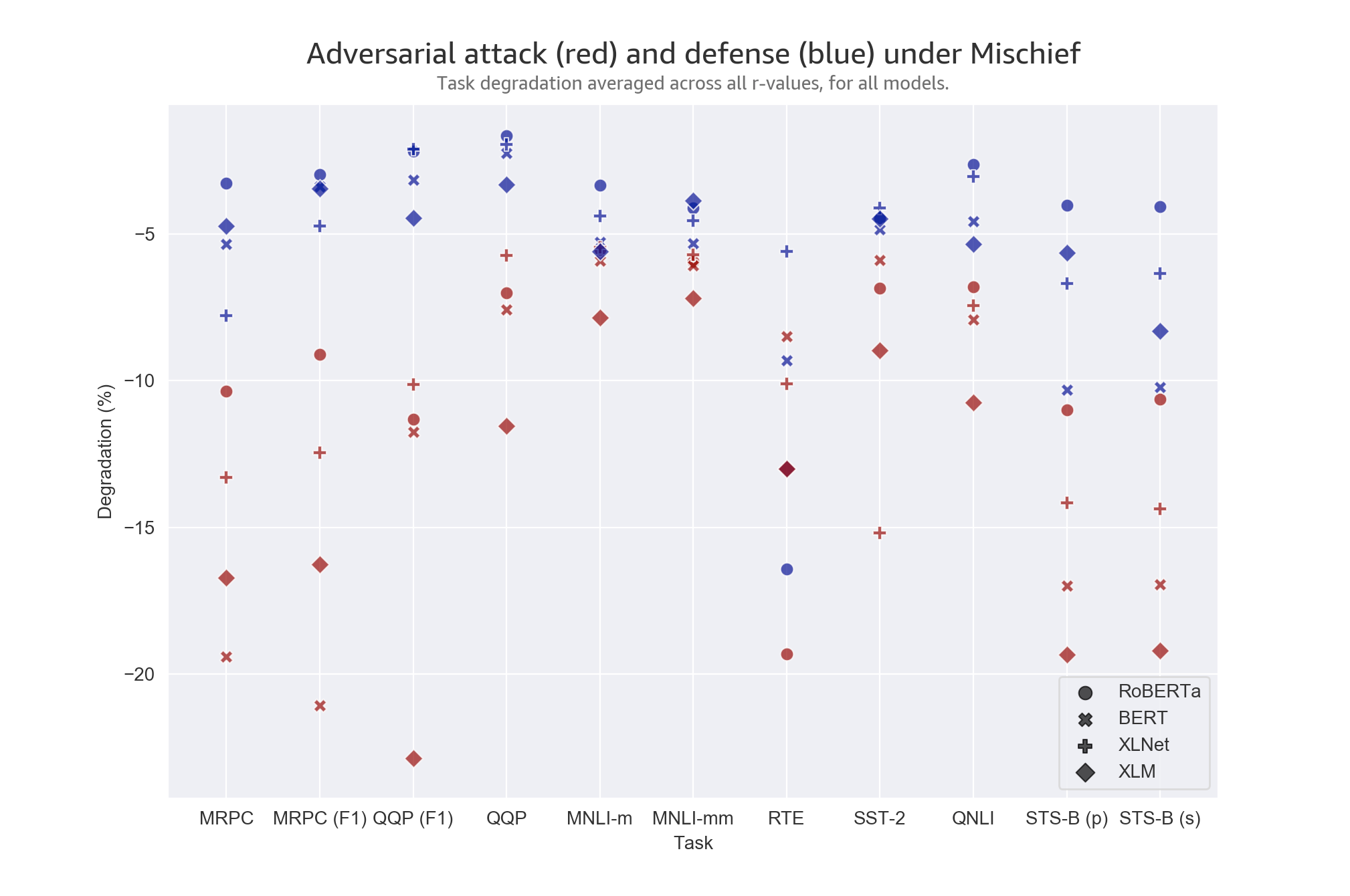}
\caption{Resulting scores under an adversarial test set, and for two situations: the adversarial setting (no Mischief on the training set, in red), and our proposed defense (with Mischief on the training set, in blue). The results are averaged out across all values of $r$, and measured in terms of F$1$, Pearson correlation ($p$ in the plot), Spearman-$\rho$ ($s$ in the plot), and accuracy (not indicated). Note the high variance in RTE, which we attribute to the small size of the dataset.}
\label{fig:onlydevset}
\end{figure}

Our first set of experiments, corresponding to the first column of \tabref{summary}, involved exploring the effects of a Mischief-generated adversarial test set, as well as a simple defense schema. To simulate an adversarial attack, on our first setup we fine-tuned the models on each GLUE task as described on their original papers, and evaluated them on test sets with Mischief. Then, we simulated a simple defense by applying Mischief to the training sets, and subsequently fine-tuning and evaluating the models. The results can be seen in \figref{onlydevset}.

\begin{figure}
\centering
\includegraphics[width=\columnwidth]{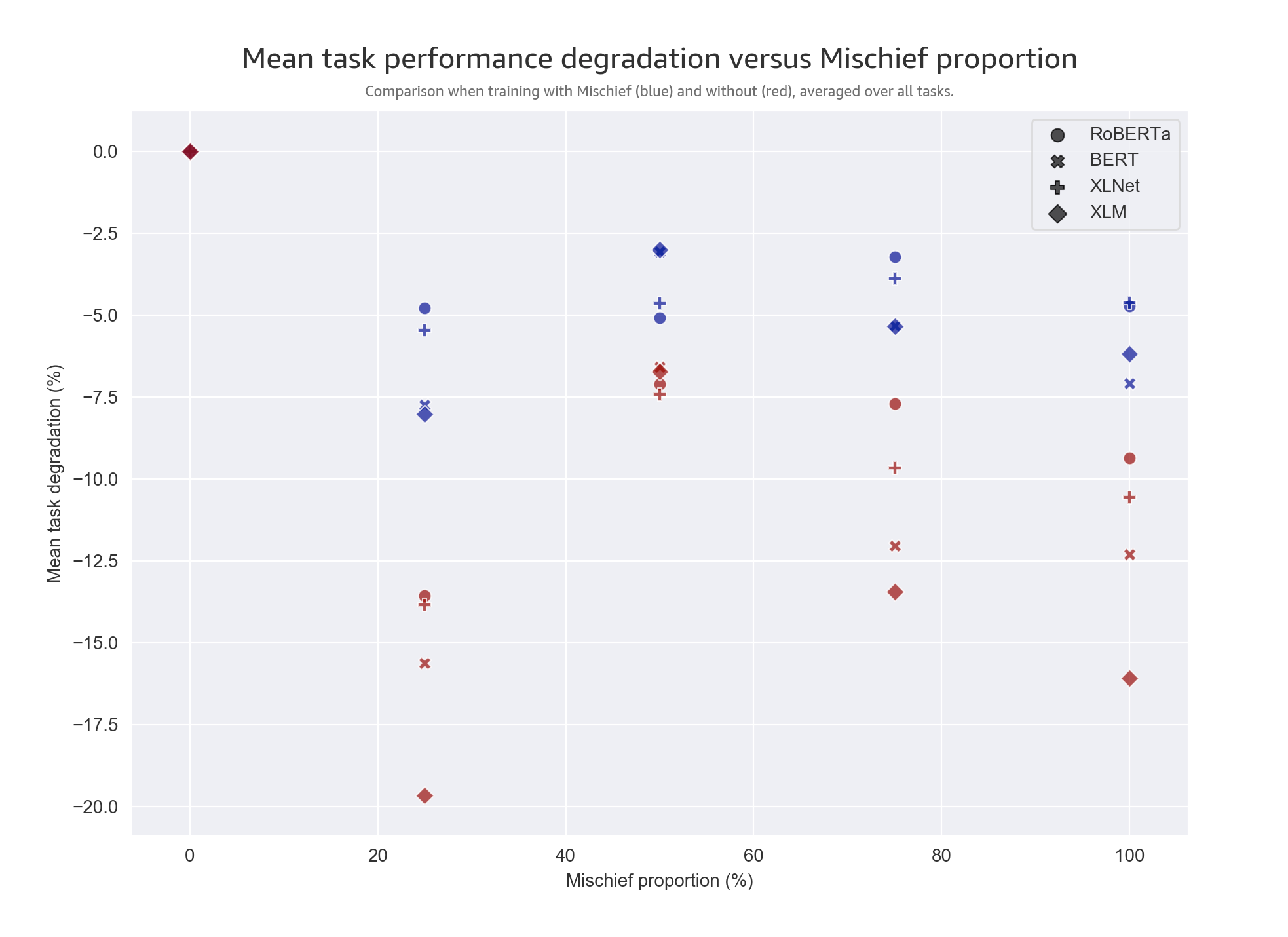}
\caption{Average (per-task) performance degradation, for varying proportions of $r$. On average, applying Mischief to the training set is an effective defense against this class of adversarial attacks. Note also how the architectures have a consistent ordering in their performance degradation, regardless of $r$.}
\label{fig:time}
\end{figure}

We found that models trained without Mischief are vulnerable to adversarial attacks, with performance drops averaging almost 20\% on the case where $r=25\%$. However, such degradation can be easily recovered by training with Mischief, as well as performing minor hyperparameter tuning to compensate for the variations in the new training set. A plot of the mean task degradation observed by varying $r$ can be seen in \figref{time}. We conjecture that the "dip" at $r=25\%$ can be explained by the fact that the concentration of GRA examples is enough to degrade the performance of the tested models, but not sufficient to allow for learning.

\subsection{Effects of Mischief in the Training Set}\label{sec:trainset}

\begin{figure}
\centering
\includegraphics[width=\columnwidth]{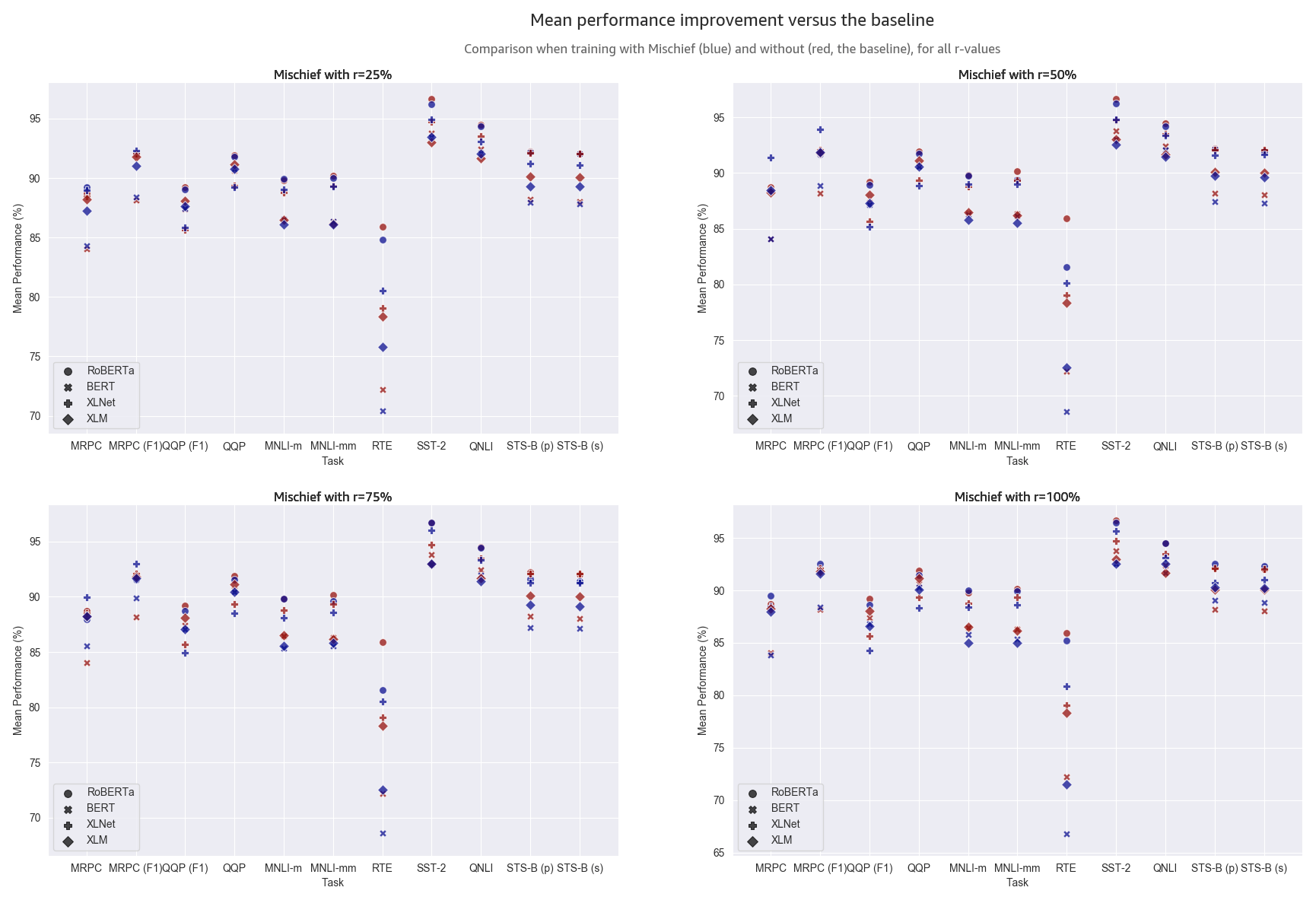}
\caption{Resulting performance change, across all tasks, for every $r$. In blue we report the best performance of a model trained with Mischief, and evaluated on its original test set. In red, we report the baseline. In general, large corpora consistently benefit from Mischief-based training.}
\label{fig:onlytrainset}
\end{figure}

Our second set of experiments involve evaluating the performance on the unmodified test dataset, after training the models with Mischief. The results for every proportion $r$ can be seen in \figref{onlytrainset}. We observed increased performances on various tasks. However, it does appear that the size of the dataset, as well as the objective of the task, have an important influence as to whether Mischief-trained models can have such performance increase. 

This is to be expected, as certain tasks do rely more heavily on "masking" certain tokens. For example, MRPC (the Microsoft Research Paraphrase Corpus, by \newcite{dolan2005automatically}), is, as it name indicates, a classification task where the model must determine whether two sentences $p,q$ are paraphrases of one another. A paraphrase of $p$ would normally retain most of the semantic content, while altering the lexical relations as much as possible, in which Mischief clearly allows for a more fine-grained data expansion at the tokenizer level. It is also important to point out that MRPC is a relatively large dataset, with $5801$ sentence pairs. 

On the other hand, some other tasks would actually be harmed by the unintentional "masking" induced by Mischief. As an example, RTE (Recognizing Textual Entailment) is a dataset merged by \newcite{GLUE} from the corpora by \newcite{DaganRTE}, \newcite{BarHaim}, \newcite{giampiccolo-etal-2007-third}, and \newcite{bentivogli2009fifth}. Its objective is to determine whether a pair of sentences $p,q$ have the relation $p \implies q$. It could be argued that such a task cannot benefit from Mischief, as it would lose critical lexical information and simply obfuscate the dataset further. However, MNLI (the Multi-Genre NLI corpus, by \newcite{williams-etal-2018-broad}) also involves textual entailment, but it is significantly larger than RTE: the latter is the smallest dataset presented in this paper, with $2769$ examples total, while the former is nearly $150$ times larger, topping about $433\!,000$ sentence pairs.

\subsection{Discussion}

Mischief as an adversarial attack is remarkably effective, although its ability to degrade the performance of a language model is, fortunately, easily lost if the model has been exposed to other GRA samples before. We hypothesize that this is, as mentioned in \secref{background}, due to the way these models construct their vocabulary. The models tested employ a byte pair encoding (BPE) preprocessing step \cite{BPE}, which segments subwords iteratively and stores them in the vocabulary based on their frequency. It follows that any model trained on Mischief-generated samples will become more robust to the perturbations induced by this algorithm. 
Moreover, the models tested have large parameter sizes, which translates into a much stronger ability to memorize, and thus be resilient to, new input examples. 

This can also help partially explain the results observed in \secref{trainset}: let $w_i, w_j$ be two words occurring in different parts of the dataset, and where $w_i = w_j$, and $|w_i| = |w_j| := n$. For $n \geq 3$, and assuming a uniform distribution, the probability that these two words are transformed the same way is 

\begin{equation}\label{eq:explanation}
\text{Pr}[\text{GRA}(w_i) = \text{GRA}(w_j)] = \left( \frac{1}{(n - 2)! } \right)^2
\end{equation}

which in turn means that Mischief is effectively a data augmentation technique--the average English word length is $n=5.1$ characters.\footnote{\url{https://www.wolframalpha.com/input/?i=average+english+word+length}} Although this number would naturally vary with the corpus being utilized, ultimately the models tested will be exposed to a wider variety of slight perturbations on the inputs, and in turn allows it to focus better on the linguistic relations between the tokens.

However, \eqref{explanation} does not account for the fact that some tasks do not benefit from Mischief. For example, the QQP (Quora Question Pairs) dataset attempts to relate a question-answer pair semantically\footnote{\url{https://www.quora.com/q/quoradata/First-Quora-Dataset-Release-Question-Pairs}}, and Mischief-based models consistently underperformed in spite of the fact that this corpus has nearly $400,\!000$ lines. Given the scores in STS-B \cite{cer-etal-2017-semeval}, and SST-2 \cite{socher-etal-2013-recursive}, it appears that, generally speaking, tasks where semantic similarity is the primary measurement are more likely to be impacted negatively. There were some exceptions to the rule, however, as some models did outperform their baseline, for example, BERT in STS-B for $r=100\%$ and XLNet in SST-2 for $r=75\%$, and $r=100\%$.

\section{Conclusion}\label{sec:conclusion}

We presented Mischief, a simple algorithm that allows us to construct a class of human-readable adversarial examples; and showed that the injection of such examples in the dataset is capable of significantly degrading the performance of transformer-based models. 
Such models can be made resistant to Mischief-based attacks by simply training with similar examples, and without relying on other components (e.g., a spell-checker).

However, Mischief has also value as a data augmentation technique, as we saw that certain NLU tasks benefit from the inclusion of such examples. 
It is important to point out that, in general, adversarial attacks are architecture-independent \cite{Szegedy}. Although we attempted to provide an in-depth analysis of select transformer-based architectures, it remains an open problem as to whether the results of this paper are applicable to other families of models. We conjecture that, as long as their tokenizer operates in a similar fashion to the WordPiece tokenizer from \newcite{wordpiece}, and their parameter size is large enough, the effects from this study extend to them. In the case of smaller-capacity models or other word-segmentation techniques where out-of-vocabulary words are frequently mapped to the same token, the outcome of a Mischief-based attack can only be more detrimental.

Finally, one area we did not pursue in this paper is synonym injection. We argue that synonym injection is arguably far more impactful in terms of supplying strong adversarial examples, and a Mischief-based approach to training with such examples may also increase performance in the tasks where Mischief did not show an improvement. However, given how sensitive is the meaning of a word--let alone their synonym--to context, such process cannot be done in an automated fashion, and without expert knowledge being invested.

\section*{Acknowledgments}
The author is grateful to B. d'Iverno, Y. Ibrahim, V. Khare, A. Mottini, and Q. Wang for their helpful comments and suggestions throughout this project, and to the anonymous reviewers whose comments greatly improved this work.

\bibliography{biblio}

\end{document}